\documentclass[10pt,twocolumn,letterpaper]{article}

\usepackage{cvpr}   
\usepackage[accsupp]{axessibility}  % Improves PDF readability for those with disabilities
\definecolor{cvprblue}{rgb}{0.21,0.49,0.74}
\usepackage[pagebackref,breaklinks,colorlinks,allcolors=cvprblue]{hyperref}

\def\paperID{CV4Clinic 2026-12} % *** Enter the Paper ID here
\def\confName{CVPR}
\def\confYear{2026}

\title{AGA3DNet: Anatomy-Guided Gaussian Priors with Multi-view xLSTM for 3D Brain MRI Subtype Classification}

\author{
Peiyu Duan$^{1}$ \quad
Xueqi Guo$^{2}$ \quad
Sepehr Farhand$^{2}$ \quad
Mehmet Berk Sahin$^{3}$ \\
Xinyuan Zheng$^{1}$ \quad
James S. Duncan$^{1}$ \quad
Gerardo Hermosillo Valadez$^{2}$ \quad
Yoshihisa Shinagawa$^{2}$ \\ [6pt]
\begin{tabular}{c c c}
$^{1}$Yale University & $^{2}$Siemens Healthineers & $^{3}$Purdue University \\
New Haven, CT, USA & Malvern, PA, USA & West Lafayette, IN, USA
\end{tabular}\\
{\tt\small \{camille.duan, xinyuan.zheng, james.duncan\}@yale.edu} \\
{\tt\small \ sahinm@purdue.edu}\\
{\tt\small \{xueqi.guo, sepehr.farhand} \\
{\tt\small gerardo.hermosillovaladez, yoshihisa.shinagawa\}@siemens-healthineers.com}
} 

\begin{document}
\maketitle
\begin{abstract}
Accurate 3D brain MRI subtype classification benefits from both localized anatomical cues and long-range contextual reasoning. We present AGA3DNet, a report-grounded framework that incorporates brief anatomical phrases extracted from radiology reports as a soft anatomical prior channel and fuses it with a lightweight 3D CNN and multi-view xLSTM aggregation. Specifically, extracted anatomical phrases are mapped to atlas-defined regions and converted into smooth spatial priors using a signed-distance transform followed by Gaussian weighting, providing interpretable, anatomy-grounded guidance without requiring dense voxel annotations. We evaluate AGA3DNet on a retrospective institutional brain MRI cohort for abnormal subtype discrimination and compare against reproducible 3D classification baselines. AGA3DNet achieves improved overall balance across performance metrics and supports clinically interpretable localization through the prior channel. We discuss limitations related to single-cohort evaluation and the lack of large-scale public brain MRI datasets paired with radiology reports under broadly usable terms.
\end{abstract}    
\section{Introduction}
\label{sec:intro}

Magnetic resonance imaging (MRI) is a cornerstone of neuroimaging, providing high-resolution structural information for computer-aided diagnosis~(CAD) of brain disorders such as tumors, white matter lesions, and multiple sclerosis~\cite{chat_diagnosis,cad_lesion,cad_ms,cad_tumor}. CAD systems aim to assist radiologists by reducing workload, improving diagnostic consistency, and enabling large-scale abnormality detection, ultimately leading to greater clinical efficiency and better patient care~\cite{cad_improve_efficiency, cad_impact,efficiency_improvement}.

Yet two challenges limit current CAD approaches. First, most methods treat MRI as the sole modality~\cite{mri_only_tumor,mri_cnn_Pereira}, while recent multimodal efforts attempt to fuse full radiology reports directly with image features \cite{huang_fusion,pellegrini_fusion,medclip}. However, directly integrating free-text reports is clinically impractical and risks diagnostic leakage, since clinical notes often contain explicit diagnostic terms, and impose an unrealistic documentation burden on clinicians~\cite{albrecht_enhancing_2025, gaffney_medical_2022}. In contrast, we propose an anatomy-aware strategy illustrated in Fig.~\ref{fig:motivation}. Instead of embedding full radiology reports, our method eliminates the need for direct text–image fusion by using reports only to extract anatomical cues (e.g., hippocampus, periventricular white matter). These cues are then spatially grounded through SynthSeg-derived~\cite{SynthSeg} structural masks and additional region annotations from MedSAM2~\cite{ma2025medsam2}, ensuring broad coverage of clinically relevant brain regions. Large language models (LLMs) serve solely as a scalable mechanism for phrase extraction, while the deployed model itself operates purely on structural MRI and anatomy-guided priors that seamlessly integrate into a CAD pipeline, thereby minimizing clinician burden and avoiding the risks associated with full-report integration.

Second, while conventional 3D convolutional networks capture local textures effectively, they struggle with long-range dependencies across distributed brain regions. Transformer-based or state-space architectures address this limitation but at a high computational cost, which hampers their practicality for large-scale deployment in clinical settings.

In summary, we present \textbf{AGA3DNet}({\textbf{A}natomy-guided \textbf{G}aussian \textbf{A}ttention}), a hybrid architecture designed for anatomy-aware CAD from brain MRI. Our main contributions are:

\begin{figure*}[t] % use [t] or [h] as you prefer
  \centering
  \includegraphics[width=0.8\textwidth]{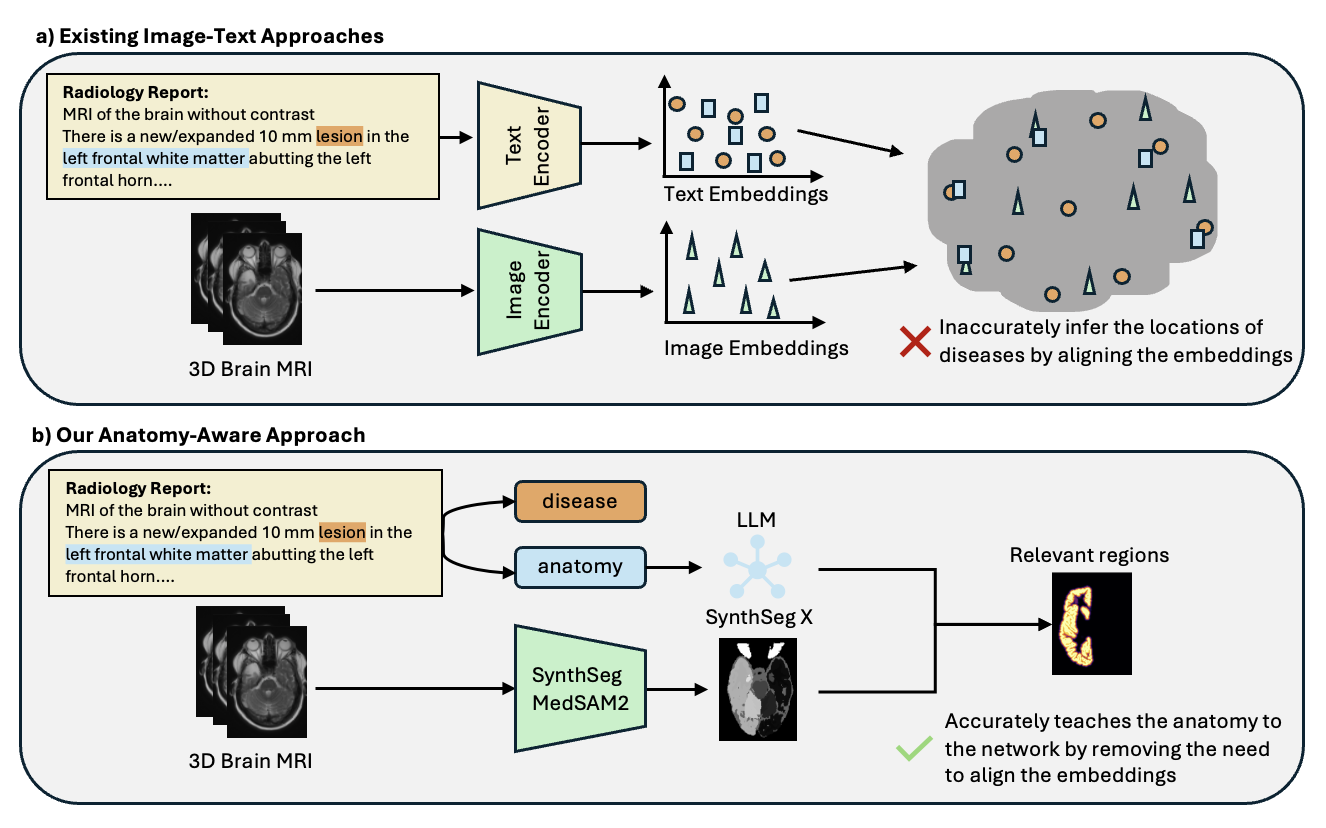}
  \caption{Comparison of anatomical abnormality detection and classification strategies for brain MRI. (a) Existing vision-language methods align global report embeddings with image features, often introducing diagnostic leakage and mislocalizing disease findings due to a lack of anatomical grounding (b) Our anatomy-aware CAD approach extracts concise anatomical phrases from reports using LLM and maps them to segmentation labels, providing more accurate and clinically practical CAD deployment.}
  \label{fig:motivation}
\end{figure*}

\begin{itemize}
    \item \textbf{Anatomy-aware CAD priors} We introduce report-grounded spatial priors by extracting short anatomical phrases with LLMs and map them onto our SynthSeg X label masks. This enriched anatomical map yields explicit coverage of both global and finer optic regions, enabling CAD systems to focus on clinically relevant regions while sparing radiologists from composing full-text reports, making the integration of textual knowledge both efficient and workflow-friendly.

    \item \textbf{Hybrid global local encoder} We design a lightweight 3D CNN for high-resolution local feature extraction, augmented with extended LSTM (xLSTM) modules that propagate information across views and volumes, enabling efficient long-range anatomical context modeling. 

    \item \textbf{Dual loss for CAD robustness} A joint focal and contrastive loss enforces discriminative embeddings while emphasizing rare and challenging cases, addressing class imbalance frequently encountered in CAD datasets.

    \item \textbf{Comprehensive evaluation} We validate \textbf{AGA3DNet} on a large 3D in-house brain MRI datasets, showing consistent improvements over CNNs, ResNet, transformer, and state-space models. These results highlight the utility of anatomically grounded report cues for enhancing disease subtype classification, indicating the potential of AGA3DNet as a clinically viable component of CAD. 
\end{itemize}
\section{Related Works}
\label{sec:related_works}

Automated medical image classification has attracted substantial attention in recent years, fueled by advances in deep neural architectures. Progress has been shaped by developments in convolutional networks, transformers, and recent state-space models, each offering complementary perspectives on spatial and temporal representation learning.

\textbf{Convolutional neural networks (CNNs)}
Deep convolutional architectures, most notably Residual Networks (ResNet)~\cite{he2016resnet}, introduced skip connections that alleviated vanishing gradients and enabled the training of very deep models. CNNs and their 2D/3D variants rapidly became the dominant backbones for brain MRI classification, offering strong inductive biases for capturing local structural patterns and hierarchical features~\cite{lecun2015deep}. Despite their success, purely convolutional designs are inherently limited in modeling long-range spatial dependencies, which are often crucial for representing distributed neuroanatomical alterations.

\textbf{Transformers in medical imaging}
Transformers~\cite{vaswani2017attention} introduced self-attention as a scalable mechanism for global context modeling. Adaptations such as Vision Transformer (ViT)~\cite{dosovitskiy2021vit} and Swin Transformer~\cite{liu2021swin} extended these ideas to image analysis, with Swin UNETR~\cite{hatamizadeh2022swinunetr} demonstrating strong performance on volumetric medical segmentation by combining hierarchical attention with UNet-style decoders. When adapted for classification, transformer backbones provide improved modeling of distributed features compared to CNNs, but their quadratic complexity can be prohibitive for high-resolution 3D scans.

\textbf{State-space sequence models}
Structured state-space models (SSMs) have recently emerged as efficient alternatives to self-attention for long-sequence modeling. Mamba~\cite{gu2024mamba} leverages selective SSMs to capture long-range dependencies with linear complexity, providing scalability advantages over transformers. Building on this foundation, MedMamba~\cite{yue2024medmamba} adapts the framework to medical imaging tasks such as MRI, CT, and histopathology, achieving competitive performance by explicitly modeling spatiotemporal context. nnMamba~\cite{gong2024nnmamba} further advances the paradigm by introducing neural parameterizations within the SSM framework, unifying convolutional feature extraction with sequence modeling. This design enables strong performance on high-dimensional medical images, making it especially promising for volumetric data. Extending beyond imaging, EHRMamba~\cite{pmlr-v259-fallahpour25a} brings the state-space approach to electronic health records, efficiently processing much longer patient histories with linear computational cost. Through multitask prompted finetuning, it supports multiple clinical tasks within a single model, improving efficiency, generalization, and real-world deployability in healthcare systems.

\textbf{Recurrent models and xLSTM}
Classical recurrent networks such as LSTMs~\cite{lstm} have historically struggled with very long contexts. Extended long short-term memory (xLSTM)~\cite{beck2024xlstm} revisits this line of work, combining LSTM gating mechanisms with efficient sequence-modeling principles inspired by SSMs. xLSTM demonstrates strong performance on long-sequence benchmarks while maintaining lower computational complexity than transformers. Its ability to balance short- and long-range dependencies makes it particularly suitable for 3D neuroimaging, where volumetric signals span multiple anatomical scales.

\textbf{Vision–language integration.}  
Recent work has explored leveraging radiology reports to enhance image-based models. For example, RGRG~\cite{rgrg}, Med-UniLM~\cite{wang2022medunilm}, and the report-guided lesion detection framework of ReportGuidedNet~\cite{lesion} demonstrate the potential of coupling textual information with imaging for report generation and abnormality classification. While these approaches improve performance, they typically use full free-text reports as global semantic signals, which risks diagnostic leakage and introduces an additional burden on clinicians. In contrast, our method focuses on extracting only concise anatomical phrases from reports and grounding them in segmentation-derived regions, enabling more practical and interpretable integration of textual knowledge.

\textbf{Positioning}
In contrast to existing CNN-, transformer-, and SSM-based approaches, our framework focuses on anatomically grounded integration of anatomy information with 3D MRI. We introduce a two-channel representation that fuses LLM-derived embeddings of report phrases with SynthSeg parcellations, enabling explicit alignment between clinical semantics and brain regions. Coupled with a lightweight 3D CNN for local feature extraction and multi-view xLSTM modules for efficient long-range modeling, our design provides a novel and interpretable pathway for robust neuroimaging analysis.
\section{Methodology}
\label{sec:method}
Our method integrates volumetric MRI data with anatomical priors derived from radiology reports. Unlike hard binarized thresholding, which would force the network to ignore all voxels outside reported regions, we generate Gaussian-weighted spatial priors. This design preserves global brain context, since regions not explicitly mentioned in the report may still carry diagnostic value, while softly emphasizing reported structures to guide the model toward clinically meaningful areas without discarding complementary information. The model receives a two-channel 3D input volume:
\begin{itemize}
    \item \textbf{Channel 1:} Conventional 3D Brain MRI volume.  
    \item \textbf{Channel 2:} A radiology-guided anatomical mask highlighting the brain regions most relevant to the clinical report.  
\end{itemize}

\begin{figure*}[htbp]
  \centering
  \includegraphics[width=\textwidth]{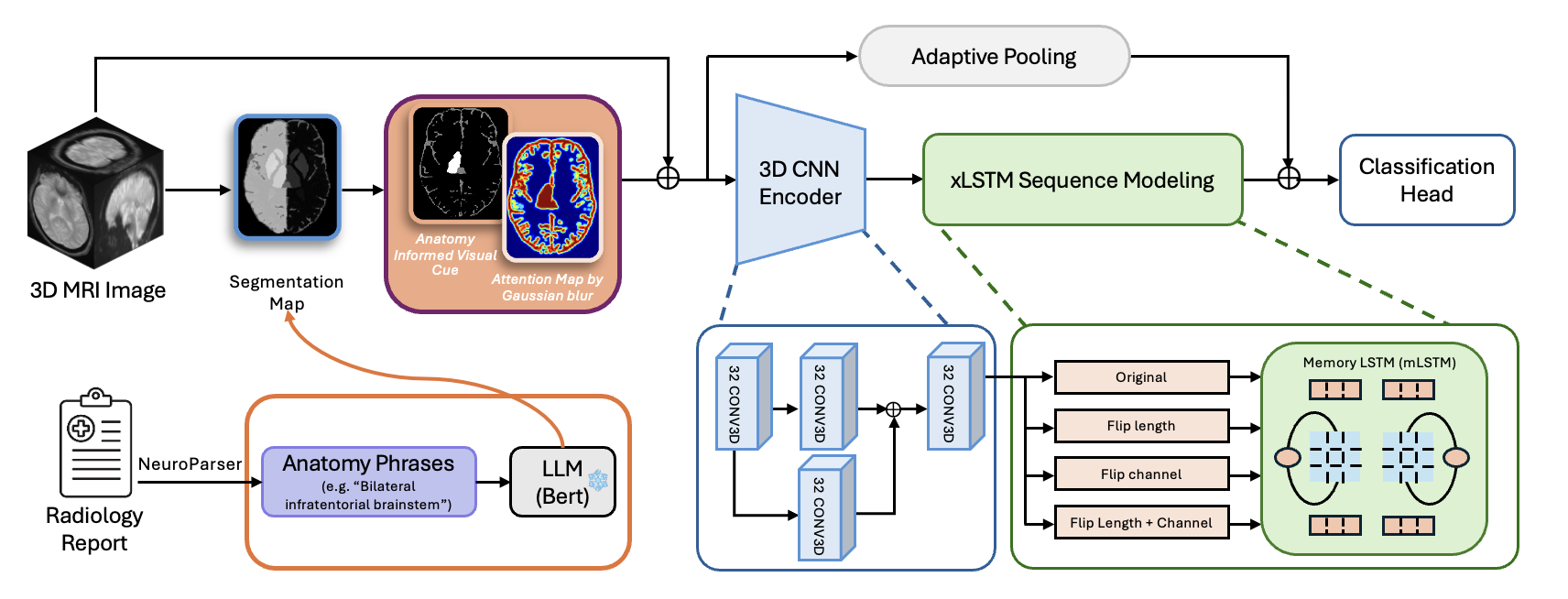}
  \caption{Schematic overview of the proposed model. The two-channel volumetric input consists of a raw T2-weighted MRI scan (channel A) and the SynthSeg Xanatomical prior (channel B). In channel B, anatomy-related phrases extracted from radiology reports are used to filter SynthSeg-derived and MedSAM2-augmented segmentation maps, yielding Gaussian-weighted salience masks that provide localized anatomical guidance. Both channels are encoded by a lightweight 3D backbone, converted into multi-view feature sequences, and integrated by mLSTM modules to produce the final classification output.
}
  \label{fig:your_label}
\end{figure*}

\subsection{Data preprocessing}

All brain MRI volumes underwent a standardized preprocessing pipeline:
\begin{itemize}
\item \textbf{Resampling:} Volumes were resampled to isotropic resolution \([256 \times 256 \times 128]\)~voxels. This ensures anatomical comparability across subjects and scanners, a prerequisite for learning consistent spatial patterns.
\item \textbf{Intensity normalization:} Each volume was normalized to zero mean and unit variance. This reduces scanner- and patient-specific intensity variations, mitigating domain shifts.
\item \textbf{Augmentation:} We applied random affine transformations (rotations, translations, scaling), elastic deformations, Gaussian noise, and intensity perturbations. These augmentations encourage robustness to positional variability and imaging artifacts, reflecting the heterogeneity of real-world clinical MRI acquisitions.
\end{itemize}

\subsection{SynthSeg X anatomical prior channel}

We refer to the added segmentation information as SynthSeg X, which is the second input channel that introduces anatomical priors by grounding concise report-derived phrases in brain structures. SynthSeg~\cite{SynthSeg}, a fast and widely used tool for whole-brain segmentation from T1/T2 MRI, is used to obtain structural delineations. A limitation of SynthSeg is its lack of coverage for certain clinically important regions; for instance, the optic nerve is absent despite its strong association with lesion burden and retinal ganglion cell loss in neuroinflammatory disease~\cite{denis2022optic,sartoretti2017optic}. To address this gap, we augment SynthSeg with additional masks generated using MedSAM2~\cite{ma2025medsam2}. Our goal is not pixel-accurate segmentation but rather coarse, consistent ROI coverage to ensure that missing but clinically relevant structures are represented, reducing the risk of systematic oversight.

To approximate the optic nerve, we implement a linear box alignment procedure. Let the reference subject have an optic nerve bounding box \(B_{\mathrm{ref}}\) with center \(c_{\mathrm{ref}}\in\mathbb{R}^3\) and side lengths \(s_{\mathrm{ref}}=(w_{\mathrm{ref}},h_{\mathrm{ref}},d_{\mathrm{ref}})\). For a target subject, let the global brain bounding box have side lengths \(g_{\mathrm{tgt}}=(W_{\mathrm{tgt}},H_{\mathrm{tgt}},D_{\mathrm{tgt}})\) and center \(c_{\mathrm{tgt}}\), with the reference global box denoted by \(g_{\mathrm{ref}}=(W_{\mathrm{ref}},H_{\mathrm{ref}},D_{\mathrm{ref}})\). We compute scaling factors $\boldsymbol{\alpha}$ for each axis 
as $(\alpha_x,\alpha_y,\alpha_z) =(\tfrac{W_{\mathrm{tgt}}}{W_{\mathrm{ref}}},\tfrac{H_{\mathrm{tgt}}}{H_{\mathrm{ref}}},\tfrac{D_{\mathrm{tgt}}}{D_{\mathrm{ref}}})$ and construct the scaling matrix $A = \operatorname{diag}(\boldsymbol{\alpha}) \in \mathbb{R}^{3\times 3}$. 

For any point $x$ $\in$ $\mathbb{R}^{3\times 3}$ (e.g., a corner or voxel coordinate of \(B_{\mathrm{ref}}\)), we apply a center-aware affine transformation
\[
T(x) = c_{\mathrm{tgt}} + A(x - c_{\mathrm{ref}}), \quad x \in \mathbb{R}^3,
\]
which rescales the reference box and re-centers it to the target subject. Applying \(T\) to the corners of \(B_{\mathrm{ref}}\) yields the estimated target box \(B_{\mathrm{tgt}}\) with new side lengths
$s_{\mathrm{tgt}} = \boldsymbol{\alpha} \odot s_{\mathrm{ref}}$
where \(\odot\) denotes element-wise multiplication. We then constrain MedSAM2 to \(B_{\mathrm{tgt}}\) to obtain a coarse optic nerve segmentation
\[
M_{\mathrm{ON}}^{\mathrm{MedSAM2}} = \mathrm{MedSAM2}\!\left(I_{\mathrm{tgt}}, B_{\mathrm{tgt}}\right),
\]
which is integrated into the multi-hot anatomical prior to form SynthSeg X.

We compare the extracted anatomical phrase embeddings against label text embeddings from SynthSeg X using cosine similarity. Formally, let $q \in \mathbb{R}^d$ denote the embedding of a report-derived phrase, and let $\mathcal{E} = \{ e_1, e_2, \dots, e_M \}, \; e_j \in \mathbb{R}^d$ be the set of anatomical label embeddings. The similarity between $q$ and each label embedding $e_j$ is computed as
\begin{equation}
s(q, e_j) = \frac{q^\top e_j}{| q |_2 | e_j |_2}
\end{equation}
We then rank $\{ s(q, e_j) \}_{j=1}^M$ in descending order and select the top-$K$ matches:
\begin{equation}
\mathcal{I}_K(q) = \operatorname*{arg\,topK}_{j \in \{1,\dots,M\}} s(q, e_j)
\end{equation}
where $K=5$ in our setting. The selected embeddings,
\begin{equation}
\mathcal{E}_K(q) = { e_j \mid j \in \mathcal{I}_K(q) },
\end{equation}
serve as radiology-guided anatomical priors that inform the second input channel of our model.

Moreover, instead of binary masks, we generate \textbf{Gaussian-smoothed prior maps} that provide soft spatial guidance. Specifically, we compute a signed distance transform for each anatomical region and convert it into a smooth, center-emphasizing weight map. Compared to hard masks, this yields a gradual transition at region boundaries and avoids forcing the network to ignore surrounding context (which remains available through the raw MRI channel). The final radiology-guided channel is obtained by incorporating these smooth prior maps, providing a spatially weighted anatomical prior for downstream modeling.
Given a binary segmentation mask \(M(x) \in \{0,1\}\) defined over spatial coordinates 
\(x \in \mathbb{R}^3\), we first compute its signed distance transform
\[
D(x) \;=\; 
\begin{cases}
\;\;\;\;\min\limits_{y \in \partial M} \|x-y\|_2, & x \notin M, \\[6pt]
-\min\limits_{y \in \partial M} \|x-y\|_2, & x \in M,
\end{cases}
\]
where \(\partial M\) denotes the boundary of the region.  
This produces a volumetric map in which negative values correspond to voxels inside the ROI and 
positive values to voxels outside.

We then convert this distance map into a smooth, spatially weighted prior. Let 
\[
d_{\mathrm{in}}(x) \;=\; \max\!\big(0,\,-D(x)\big),
\]
which measures the (non-negative) distance of a voxel inside the ROI to the boundary, and is zero outside the ROI. We define the Gaussian-smoothed prior as
\[
W(x) \;=\; 1 - \exp\!\left(-\frac{d_{\mathrm{in}}(x)^2}{2\sigma^2}\right),
\]
where \(\sigma\) controls the degree of smoothing. Unlike a binary mask, \(W(x)\in[0,1)\) varies smoothly within the ROI, assigning larger weights to voxels deeper inside the anatomical region and smaller weights near the boundary. This produces a soft prior channel that encourages the model to focus on clinically salient regions without discarding global context available from the raw MRI.

\begin{figure*}[htbp]
  \centering
  \includegraphics[width=0.7\textwidth]{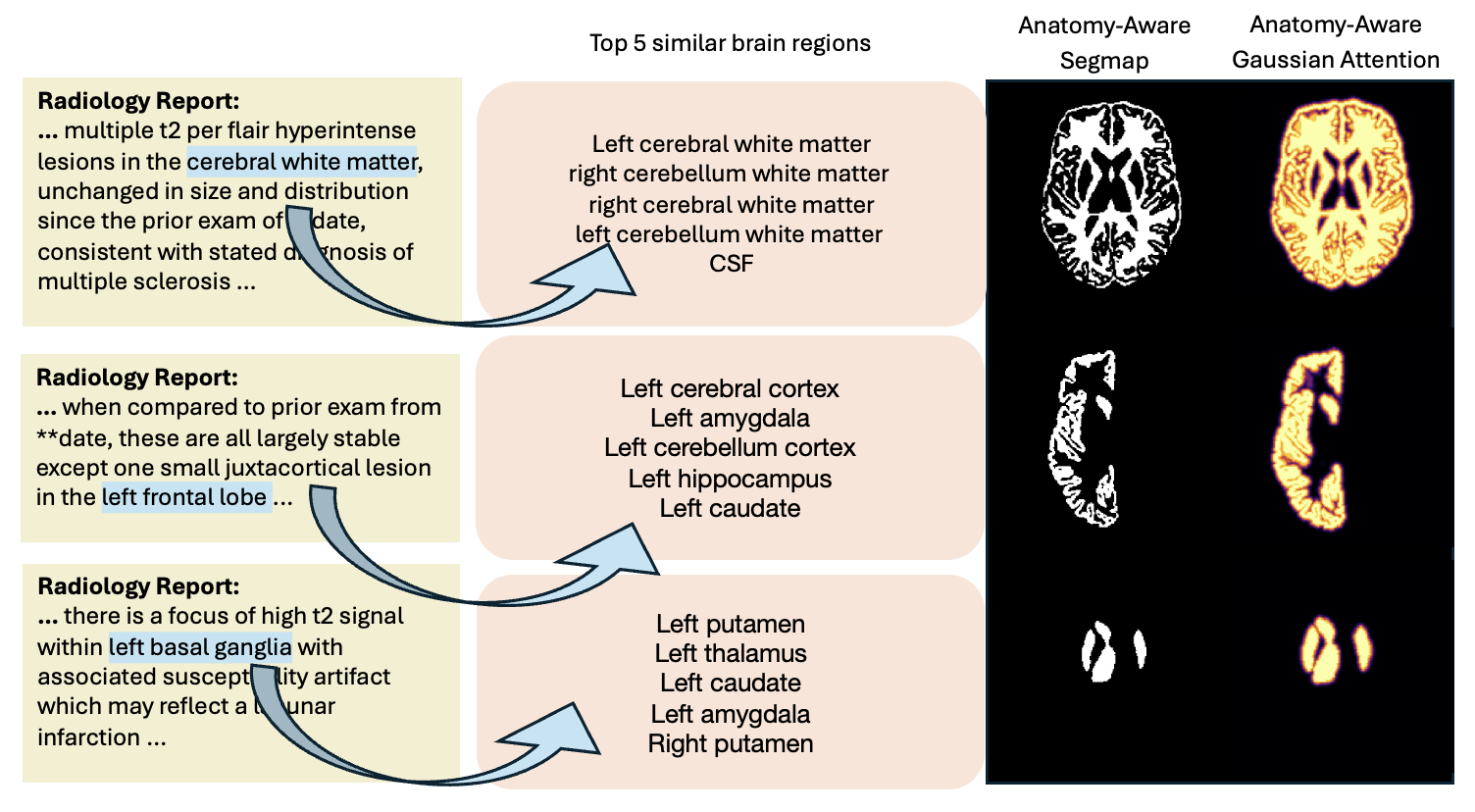}
  \caption{
  Overview of our report-guided anatomy alignment examples. Each row shows representative radiology report excerpts, the top-5 matching anatomical phrases/labels retrieved from SynthSeg X, the corresponding anatomy-aware segmentation map, and the Gaussian attention map. 
}
  \label{fig:anatomy}
\end{figure*}

This process is illustrated in \cref{fig:anatomy}, where radiology-derived phrases are matched to SynthSeg~X labels and the corresponding regions are visualized with Gaussian-blurred attention maps. The figure highlights how textual cues are grounded in anatomical structures and transformed into smooth, spatially weighted priors before model inference.

\subsection{Volumetric convolutional backbone}

Our imaging backbone is a compact 3D convolutional feature extractor tailored for volumetric brain MRI. It comprises three convolutional layers, each with 32 channels, and incorporates an intermediate residual block: the output of the first layer is routed through this residual pathway and then fused elementwise with the activations of the second layer.  

This design offers several advantages. The residual connection mitigates vanishing gradients and stabilizes training, which is particularly important given the limited size of neuroimaging datasets. The shallow depth and fixed-width 32-channel blocks reduce computational burden compared to deep volumetric CNNs, yielding a memory-efficient architecture. At the same time, the hierarchical progression from low-level filters to more abstract representations enables the encoder to capture both local tissue contrasts and broader volumetric context. Unlike transformer-based backbones, which achieve strong global modeling but incur quadratic complexity, our design provides a lightweight alternative that preserves discriminative power while remaining scalable to high-resolution 3D scans.  

By avoiding excessively deep stacks while maintaining consistent channel capacity, the backbone offers an efficient and robust foundation for subsequent sequence modeling with mLSTM modules, which explicitly capture long-range dependencies across axial, coronal, sagittal, and volumetric views.

\subsection{Multi-View xLSTM Modeling}

To capture long-range spatial dependencies, we reshape the encoder feature maps into multiple sequence representations. After the 3D convolutional backbone, each feature map has dimension $32 \times 8 \times 8 \times 4$ (32 channels, $8{\times}8{\times}4$ spatial grid). From this tensor we form four parallel sequences: (i) axial slices of length $8$, (ii) coronal slices of length $8$, (iii) sagittal slices of length $4$, and (iv) the full 3D volume flattened into a sequence of length $8*8*4$. Each element in a sequence corresponds to a 32-dimensional feature vector with 512 tokens.  

Every sequence is processed with an extended long short-term memory (xLSTM) module~\cite{beck2024xlstm}, configured with a hidden size of 2 layers. The xLSTM augments standard gating with cross-dimensional interactions, enabling the model to propagate information across distant anatomical locations. The outputs from the axial, coronal, sagittal, and volumetric branches are then aggregated via average pooling into a unified embedding of dimension 512. This embedding is passed through a lightweight classifier (two fully connected layers with ReLU and sigmoid output) to predict the disease subtype.  

This hybrid design combines efficient local feature extraction from the 3D CNN with global context modeling from mLSTMs, while keeping the parameter count and memory footprint manageable. Explicitly specifying the sequence dimensions, hidden sizes, and fusion strategy facilitates reproducibility and highlights the balance between efficiency and discriminative power.

\subsection{Classification head}

The fused embedding is passed through two fully connected layers with ReLU activations, followed by a final prediction layer. We employ a sigmoid activation for the final class prediction.

\subsection{Training objectives}

We employ a hybrid objective that couples contrastive learning with focal loss.  
To address the strong class imbalance in our datasets, we adopt the focal loss~\cite{lin2017focal}, which reshapes the standard cross-entropy loss to focus training on harder, misclassified examples:

\begin{equation}
\label{eq:focal}
\mathcal{L}_{\text{focal}}(p_t) \;=\; - \alpha \, (1 - p_t)^{\gamma} \, \log(p_t),
\end{equation}

where $p_t$ denotes the predicted probability of the ground-truth class, 
$\alpha \in [0,1]$ is a weighting factor that balances positive and negative samples, 
and $\gamma \geq 0$ is a focusing parameter that reduces the relative loss for well-classified examples. 
Both hyperparameters are selected based on validation performance.

The term $(1 - p_t)^{\gamma}$ down-weights well-classified examples (those with large $p_t$), preventing them from dominating the loss. For underrepresented or harder classes, where $p_t$ tends to be smaller, the modulation factor is close to one, so their contribution to the loss is preserved. This property makes focal loss especially beneficial in imbalanced medical imaging tasks, as it encourages the model to focus on subtle or rare pathological patterns rather than being biased toward the majority class.

Given a batch of $N$ normalized embeddings $\{\mathbf{z}_i\}_{i=1}^N$ with labels $\{y_i\}_{i=1}^N$, 
The supervised InfoNCE loss for anchor $i$ is defined as \begin{equation}
\mathcal{L}_{\text{SupCon}} 
= -\sum_{i=1}^N \frac{1}{N|\mathcal{P}(i)|} 
\sum_{p \in \mathcal{P}(i)} 
\log\frac{\exp\left(\mathbf{z}_i^\top \mathbf{z}_p / \tau\right)}
{\sum_{a \neq i}\exp\left(\mathbf{z}_i^\top \mathbf{z}_a / \tau\right)}
\end{equation}
where $\mathcal{P}(i)$ is the set of positives ($y_p = y_i,\, p \neq i$) and $\tau$ is a temperature. 
This loss encourages embeddings of samples with the same label to cluster 
while separating those with different labels.

This objective encourages embeddings from the mLSTM branches to cluster within classes while remaining well-separated across classes, thereby improving discriminability in challenging settings.

To address class imbalance, we combine contrastive learning with focal loss:

\begin{equation}
\mathcal{L} = \lambda \mathcal{L}_{\text{SupCon}} + \mathcal{L}_{\text{focal}},
\end{equation}

where $\mathcal{L}_{\text{focal}}$ down-weights easy negatives and emphasizes harder examples. The trade-off coefficients $\lambda$ are tuned via validation.

Optimization is performed with Adam (initial learning rate $1 \times 10^{-4}$), cosine annealing, batch size of 4, and 100 training epochs on NVIDIA A100 GPUs. Early stopping is triggered once validation performance plateaus.
\section{Experimental analysis}

\begin{table*}[ht!]
\centering
\caption{Performance comparison of different models. Best values per column are in bold.}
\label{tab:combined_models}
\resizebox{0.62\linewidth}{!}{
\begin{tabular}{llccccc}
\toprule
&
 \textbf{Model} & \textbf{Acc} & \textbf{AUC} & \textbf{Precision} & \textbf{Recall} & \textbf{F1 Macro} \\
% \midrule
% &
% ResNet~\cite{he2016resnet} & \textbf{0.9304} & 0.9246 & \textbf{0.9324} & 0.9928 & \textbf{0.7912} \\
% & MedMamba~\cite{yue2024medmamba} & 0.8797 & 0.5000 & 0.8797 & \textbf{1.0000} & 0.4680 \\
% & nnMamba~\cite{gong2024nnmamba} & 0.8797 & 0.5000 & 0.8797 & \textbf{1.0000} & 0.4680 \\
% & Swin-UNETR~\cite{tang2022swinunetr} & 0.9051 & 0.7921 & 0.9079 & 0.9978 & 0.6742 \\
% & ResNet--mLSTM & 0.8987 & 0.8114 & 0.9073 & 0.9856 & 0.6647 \\
% & \textbf{Ours (AGA3DNet)} & \textbf{0.9304} & \textbf{0.9406} & 0.9262 & 0.9928 & 0.7649 \\
\midrule
& ResNet~\cite{he2016resnet} & 0.5631 & 0.5775 & 0.8174 & 0.5767 & 0.5023 \\
& MedMamba~\cite{yue2024medmamba} & 0.7910 & 0.5000 & 0.7913 & \textbf{1.0000} & 0.4417 \\
& nnMamba~\cite{gong2024nnmamba} & 0.7910 & 0.5000 & 0.7913 & \textbf{1.0000} & 0.4417 \\
& Swin-UNETR~\cite{tang2022swinunetr} & 0.8200 & 0.6640 & 0.8204 & 0.9816 & 0.6100 \\
& ResNet--mLSTM & 0.7860 & 0.6971 & 0.7902 & \textbf{1.0000} & 0.4402 \\
& \textbf{Ours (AGA3DNet)} & \textbf{0.8250} & \textbf{0.7236} & \textbf{0.8290} & 0.9877 & \textbf{0.6280} \\
\bottomrule
\end{tabular}}
\end{table*}
\textbf{Dataset}  
This dataset contains 216 patients with 581 scans, yielding 923 image--anatomy phrase pairs extracted using the same pipeline described in Sec.~\ref{sec:method}. 3D T2-weighted and FLAIR sequences were acquired. Data were partitioned at the patient level, to prevent data-leakage, into training (70\%), validation (15\%), and testing (15\%) subsets. Scans with lesions, demyelination, and focal dysplasia form the positive class, and all other subtypes form the negative class. This task requires discrimination among pathological subtypes with overlapping structural patterns, making it clinically relevant and challenging.

\subsection{Disease subtype classification result}

We evaluated the performances in distinguishing lesion and demyelination from other brain abnormalities, which requires discrimination among pathological subtypes with highly overlapping structural patterns, making it both more challenging and clinically relevant. Results are summarized in \cref{tab:combined_models}.

Swin UNETR achieves very high recall (0.9816), but this comes at the cost of macro-F1 (0.6100), indicating limited discriminative balance. Resnet only achieved 56.31\% accuracy, highlighting its limited generalization on disease subtypes. MedMamba and nnMamba both achieve perfect recall (1.0), but suffer from very low AUC and F1 macro, suggesting they over-predict the positive class. In contrast, our proposed AGA3DNet achieves the best AUC (0.7236) while maintaining competitive recall (0.9877), underscoring that it remains effective in more challenging subtype discrimination.  This demonstrates that our framework provides the most reliable balance between sensitivity and specificity.

\subsection{Ablation studies of disease subtype classification}
\begin{table*}[ht!]
\centering
\caption{Ablation study of our method on disease subtype classification. \checkmark indicates the presence of a component. Best values per metric are highlighted in bold.}
\label{tab:datasetB_ablation}
\resizebox{0.7\linewidth}{!}{
\begin{tabular}{lccccccccc}
\toprule 
\textbf{BCE} & 
\textbf{Focal} & \textbf{Contrastive}  & \textbf{Gaussian} & \textbf{Acc} & \textbf{AUC} & \textbf{Precision} & \textbf{F1 Macro} & \textbf{Recall}\\
\midrule
-- & \checkmark & -- & -- &0.7961 & 0.6445 & 0.8270 & 0.6115 & 0.9387\\
\checkmark & -- & -- & \checkmark & 0.801 & 0.6934 & 0.8081 & 0.5413 & 0.9816\\
--&  \checkmark& -- & \checkmark & 0.81 & 0.6412 & 0.8131 & 0.5636 & 0.9877\\
% -- & \checkmark & \checkmark & -- & 76.21 & 0.6385 & 0.8167 & 0.5735 & 0.9018\\
--&  \checkmark  & \checkmark  & \checkmark & \textbf{0.825} & \textbf{0.7236} & \textbf{0.8290} & \textbf{0.6280} & \textbf{0.9877}\\
\bottomrule
\end{tabular}}
\end{table*}

We performed ablation studies to disentangle the contributions of each component, with results summarized in Table~\ref{tab:datasetB_ablation}. These analyses confirm the validity of our design:
(1) Gaussian salience masks provide anatomy-aware spatial guidance that improves attention to clinically relevant regions;
(2) focal loss mitigates class imbalance, improving sensitivity to minority subtypes; and
(3) contrastive supervision stabilizes the effect of Gaussian priors, yielding more separable and clinically interpretable feature spaces.

\textbf{Focal loss without Gaussian prior.}
We evaluated the model with focal loss but without Gaussian weighting (row~1), using this setting to establish a prior for subsequent variants. This configuration achieved an accuracy of 79.61\%, AUC of 0.6445, precision of 0.8270, macro-F1 of 0.6115, and recall of 0.9387. 

\textbf{Gaussian prior with binary cross entropy (BCE) Loss}
Using report-guided Gaussian-weighted anatomical masks with a standard BCE loss (row~1) produces 80.1\% accuracy, 0.6934 AUC, 0.8081 precision, 0.5413 macro-F1, and 0.9816 recall. The strong recall shows that the model leverages anatomy-aware priors to localize disease-related regions. However, the relatively low AUC and macro-F1 indicate that training solely with BCE makes the model overly sensitive to salience cues, limiting general discriminability.

\textbf{Focal loss and Gaussian prior}
Training the network with Focal Loss and report-guided Gaussian-weighted anatomical priors (row~3) increases sensitivity (recall = {0.9877}) and yields balanced accuracy (81.0\%) and precision (0.8131). Although AUC decreases slightly (0.6412), this configuration demonstrates the synergy between focal loss and anatomy-informed guidance, confirming that Synthseg X channel effectively constrains attention to relevant structures without requiring full report alignment.

\textbf{Focal and contrastive with Gaussian prior}
Our full model (row~4), integrating focal loss, contrastive supervision, and Gaussian priors, achieves the best overall performance: \textbf{82.5\%} accuracy, \textbf{0.7236} AUC, \textbf{0.8290} precision, \textbf{0.6280} macro-F1, and \textbf{0.9877} recall. The contrastive term enforces greater inter-class separation in the embedding space, counteracting the over-emphasis of Gaussian priors on positive regions. Meanwhile, focal loss preserves sensitivity to rare classes. Together, these components yield the strongest combination of discriminability, robustness, and clinical interpretability.

\section{Discussion and Conclusions}

Our study demonstrates that explicitly grounding structural MRI analysis in anatomical priors extracted from radiology reports improves 3D brain disease classification. By combining lightweight 3D convolutional encoders with multi-view mLSTM modules and a joint contrastive–focal objective, \textbf{AGA3DNet} achieves higher accuracy and stronger generalization than state-of-the-art CNN, transformer, and state-space baselines. Crucially, the anatomy-aware channel enhances interpretability by linking anatomical cues to localized regions in the brain, providing model decisions that align more naturally with clinical reasoning. This design makes the system not only more accurate but also more actionable as part of CAD. In practice, our framework reduces clinician burdens: instead of requiring full free-text reports at inference, the model relies only on concise anatomical phrase cues, making integration into routine workflows more feasible.
Comparing to existing MRI foundation models \cite{brainfound} \cite{decipher}, which mainly encodes global anatomical structure or transferable representations, our method incorporates an explicit patient-specific abnormality prior through a 3D binary mask, enabling the classifier to directly exploit localized pathological evidence.

\textbf{Limitations} The quality of extracted report phrases can be variable, and coverage of finer regions remains limited. These challenges present opportunities for improvement through more precise clinician-annotated anatomy labels and expanded segmentation priors.

\textbf{Future directions} We identify several promising avenues. First, using anatomical guidance from radiology reports instead of full diagnostic narratives creates a pathway to general anatomical priors that can support a wide range of CAD tasks, such as the detection of multiple disorders and treatment response assessment, without increasing clinician workload. Second, advances in natural language processing, such as domain-adapted large language models, may improve the precision and robustness of anatomical phrase extraction across varied reporting styles. Third, adaptive or hierarchical strategies for selecting regions of interest could allow the network to dynamically prioritize relevant anatomical areas depending on patient context, instead of relying on a fixed $k$. Finally, progress toward deployment will require improving trustworthiness, including reliable uncertainty estimation, calibration, and validation across multiple imaging sites, and evaluation across diverse patient groups. Taken together, these directions highlight the potential of anatomy-guided MRI classification as a scalable and clinically practical foundation for future CAD systems. 
{
    \small
    \bibliographystyle{ieeenat_fullname}
    \bibliography{main}
}


\end{document}